# Self-Supervised Learning based Depth Estimation from Monocular Images

A. Mishra, M. Poddar, M. Kewlani
Mentor: H. Pei

*Abstract-* **Depth Estimation has wide reaching applications in the field of Computer vision such as target tracking, augmented reality, and self-driving cars. The goal of Monocular Depth Estimation is to predict the depth map, given a 2D monocular RGB image as input. The traditional depth estimation methods are based on depth cues and used concepts like epipolar geometry. With the evolution of Convolutional Neural Networks, depth estimation has undergone tremendous strides.**

**In this project, our aim is to explore possible extensions to existing SoTA Deep Learning based Depth Estimation Models and to see whether performance metrics could be further improved. In a broader sense, we are looking at the possibility of implementing Pose Estimation, Efficient Sub-Pixel Convolution Interpolation, Semantic Segmentation Estimation techniques to further enhance our proposed architecture and to provide fine-grained and more globally coherent depth map predictions. We also plan to do away with camera intrinsic parameters during training and apply weather augmentations to further generalize our model.**

## I. INTRODUCTION

The main bottleneck for the entire field of Artificial intelligence: labels, even in a field with ample data the arduous tasks of labeling them limits the progress, and in a way makes the deep learning/machine learning networks merely function mapper connecting a certain point in input to a certain point in the label. Recently, there has been an increase in the interest to get rid of this bottleneck with the development of self-supervised learning, which initially prompted to include a set of positive and negative samples to learn from a dataset.

Our task is that we want to infer a dense depth image from a single color input image automatically. Without a second input image to enable triangulation, estimating absolute or even relative depth seems ill-posed. Humans, on the other hand, learn through navigating and interacting in the real world, allowing us to make credible depth estimates for new scenarios.

Many application domains, such as autonomous driving, require good depth perception. Because of their precision and reliability, such systems have traditionally relied on data fusion from depth sensors such as LiDAR. However, such sensors are limited in their applicability under extreme weather circumstances (fog or heavy rain), have a short range, are more expensive, and are more complicated. Some of these drawbacks can be mitigated in part by estimating depth from RGB photos. While supervised learning approaches have been shown to estimate depth without the use of sensors during inference, ground truth supervision is still required to complete the training. However, gathering a large and diverse dataset is a difficult undertaking. Recent self-supervised approaches have proven that monocular depth estimation models can be trained using only stereo pairs or monocular video.

We intend to implement Monodepth2's baseline model [6] specifically on only monocular video frames, as well as various benchmark architectures, compare the performance of each model using various measures, and select the optimal model for pose estimation. We'll make the camera parameters learnable for pose estimation and compare them to existing self-supervised models. And see whether our method can help us enhance the baseline model. All the experiments are performed on the KITTI [24] dataset.

## II. OVERVIEW AND PROJECT ACCOMPLISHMENT

*A. Literature Review - A. Mishra, M. Poddar, M. Kewlani*

With the advent of autonomous systems, depth estimation has become one of the key areas to work on. Keeping commercial usage in mind, we need to think towards reducing the overall cost associated with this area. Various research in monocular and stereo depth estimation has been done in recent years, thereby reducing the dependency on LiDar hardware.

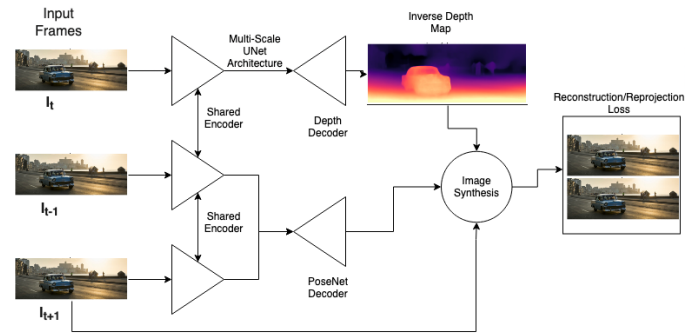

Fig. 1 Monodepth Architecture with monocular set of image frames

MonoDepth2: Digging into self-supervised monocular depth estimation

While performing research about depth estimation, we came across monodepth2 paper [6] which is a self-supervised implementation of depth estimation. On further review of this, the implementation was deemed as an ideal base implementation for our project. Monodepth paper used a stereo pair of images for depth estimation. We constructed one of these images from another image using the proposed network and calculated reconstruction loss for further reducing the loss. MonoDepth2 [6] adds upon the same idea as the original paper but implements the concept of a given set of consecutive frames in a self-supervised manner. Here, two encoder-decoder networks are being used, one for depth estimation and other for pose estimation. We use a shared encoder for pose network and depth network, while keeping different decoders for each of these. Finally we compute disparity loss in the depth network and Reprojection error for the final output. Reprojection loss depicts how well we have located the same pixel/object from initial frame to the other frame.

3D Ken Burns Effect from a Single Image: Adding semantic segmentation using Mask R-CNN:

We have chalked out a few architectural changes which might give us further improvement on the current model. To avoid semantic distortions that we see in the current model, we aim to adjust the disparity map using Mask R-CNN segmentation. This assures us that the saliency feature map is aligned to a coherent plan. Further details can be reviewed from the paper proposed by folks from Portland University and Adobe research team [19].

CamLessMonoDepth[10]: Monocular Depth Estimation with Unknown Camera Parameters

The paper proposes a self-supervised monocular technique to learn depth for pinhole cameras with minimal distortion, without providing camera intrinsics. There are a variety of methods for estimating camera parameters, including geometric structures, calibration targets, and independent neural networks. However, these methods have the disadvantage of requiring additional data for calibration, as well as increased complexity and training time. As a result, this research proposed an architecture that included a pose network and a camera network for learning camera settings from input video sequences. The sequential frames are fed into the pose network, which produces a rigid transformation with 6 degrees of freedom (DOF) from the target image plane, comprising rotational and transitional information.This is fed into the camera network, which produces a matrix of camera parameters called K. They then recreate the target image in 3D using the estimated depth and K from the input image. To approximate camera parameters, they use the photometric error metric. The paper outperforms the other state-of-the-art self supervised monocular depth estimation models. We intend to implement this approach on our proposed model and compare our models to state-of-the-art models.

ESPCN: Real-Time Single Image and Video Super-Resolution Using an Efficient Sub-Pixel Convolutional Neural Network [9]

The authors propose a Sub-Pixel Convolutional Layer for Image and Video Super-Resolution. In a nutshell, what a Sub-Pixel Convolutional Layer does is it takes a Low Resolution Image and convolves it with a fractional stride of 1/r followed by a convolutional layer with stride 1. The model computational complexity, thus, increases by a factor of r^2. To mitigate this, ESPCN dictates that weights between pixels not be calculated [9] and thus makes the computation faster. Inorder to generate more coherent depth map, we plan to leverage Efficient Sub-Pixel Convolutional Layer [9] to take a Low Resolution to High Resolution approach for depth maps instead of the general interpolation techniques for consistency. Since Sub-Pixel Convolutional Layers have done wonders for Image Reconstruction, a similar approach for a depth map could improve the depth estimation accuracies.

ConvNeXt : A ConvNet for the 2020s [18]

The paper in a nutshell intends to address the dominance of ConvNets in computer vision tasks while achieving similar performance metrics as Vision Transformers and Swin Transformers. The authors focus on the training recipe adopted by Vision Transformers like using AdamW optimizer, improved data augmentation techniques, inverted bottleneck convolutions, large kernel sizes, enhanced activations with GeLU instead of traditional ReLU, Layer Normalisation instead of BatchNorm. The observations described by the author make ConvNeXt comparable or in some instances exceed the performance metrics of transformers while at the same time reducing the floating point operations and using significantly less memory. We intend to experiment with this architecture for self-supervised depth estimation training.

*B. Implementing Monodepth2 - M. Poddar, M. Kewlani*

We adapt the basic implementation of Monodepth2 from the official repository [11]. The problem statement here is to take non-stereo monocular images and predict a disparity map of the image. The disparity map is then used along with camera intrinsics to generate the depth map. The implementation is coded in pytorch. We start by preparing the KITTI[24] dataset which has monocular images frame-by-frame and velodyne LiDar points along with camera intrinsics. The images are first resized to 192 X 640 pixels. They are then down-scaled to a factor of 2, 4 and 8 for reconstruction loss assimilation. Additionally, we augment the images with random horizontal flip (p=0.5) and color jitter (p=0.5). The images are then encoded to tensors.

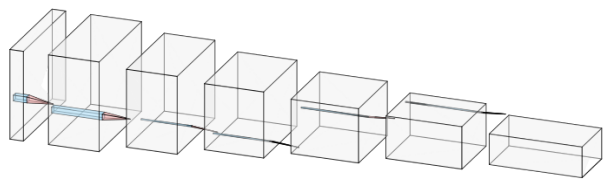

Fig. 2 ResNet50 architecture pretrained model on ImageNet used as backbone (encoder). The linear layers followed by convolutions aren't used from the architecture.

The next crucial part of implementation is encoding the image. We use ResNet-50 model for encoder with pretrained (on ImageNet) weights of initial convolution layer, batchnorm layer, and resnet layers 1 through 4. We don't use the linear layers of the architecture as we don't need to classify the images. During the forward operation of the encoder model, we keep the outputs of layers 1 through 4 which correspond to the scales.

Next, we move to the Depth Decoder which is a 4 step "up convolution". This essentially is passing the input to a convolutional block with filters 3X3 and upsampling it by a factor of 2 using interpolation with mode nearest. At each upsample step, we concatenate the output features of the encoder model at various scales. This is done to mimic the architecture of UNet.

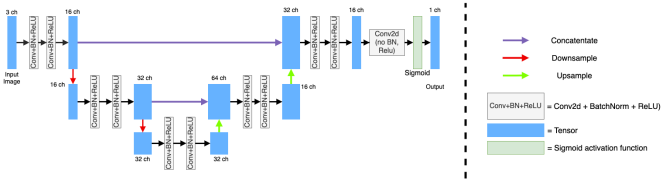

Fig. 3 Sample UNet Architecture taken from ECE-GY 6123 Computer Assignment 4.

The output of the Depth Decoder is a 1 channel image which is the inverse depth map. Using the inverse depth map / scaled disparity map, we calculate the scaled depth map. Now, since we are using self-supervised training, we don't essentially have the depth ground truth. The authors suggest that since we have the -1 and +1 frames, we use a Pose Estimation network.

The Pose Estimation network takes encoded features of two frame pairs [-1, 0] or [0, 1] using the same shared ResNet encoder. It performs squeeze convolution followed by 3 convolutional blocks, generates its mean and finally calculates its axis angle and translation vectors between the input frame and target frame. The axis angle is then used to calculate the rotation of the target frame (-1 or 1) with respect to the 0th frame, which is then combined with translation to generate an "essential" matrix R't ($T_{t \to t'}$).

The scaled depth map $D_t$, $T_{t \to t'}$ and the pseudo inverse of camera intrinsics matrix (K) are then used to backproject the depth. The backprojection operation is simply a matrix multiplication of the scaled depth values to the inverse K matrix over x and y axes. Finally, the back projected depth points of the 0th frame are warped on the -1 and +1 frames. This is done by first multiplying the K matrix with $T_{t \to t'}$. The output is then multiplied with the outputs of depth back projected points of the 0th frame at each scale. This is essentially the flow-field grid from the depth points.

The last operation is to generate color images of -1 and +1 frames using the original color -1 and +1 images and flow-field grid using grid_sample operation. The final predicted output -1 and +1 images are then used as a substitute to calculate the losses.

*C. Run and Performance Metrics Evaluation - A. Mishra, M. Kewlani*

We run the implemented monodepth2 architecture on NYU's HPC clusters with 14 cpu cores, 128 GB memory, Nvidia RTX8000 GPU - 45 GB memory. To perform the run, first we need to formulate the losses.

Computing loss is an integral part of training the neural network. In a self-supervised learning approach, the loss calculation seems dubious. There are a number of losses which we use to generate the combined loss to train. First, we use reprojection loss across each scale using the mean absolute difference. Here, α = 0.85.

$$pe(I_a, I_b) = \frac{\alpha}{2}(1 - SSIM(I_a, I_b)) + (1-\alpha)||I_a - I_b||_1$$

$$I_{t' \to t} = I_{t'} < proj(D_t, T_{t \to t'}, K) >$$

$$L_p = min(\sum pe(I_t, I_{t'}))$$

Next, we use the inverse depth map's gradients in x and y directions, and the 0th frame's gradients in x and y directions. The two are combined using the below equation to calculate the smoothing loss.

$$L_s = |\partial_x d_t^*|e^{-|\partial_x I_t|} + |\partial_y d_t^*|e^{-|\partial_y I_t|}$$

The final loss is calculated using a linear combination of Lp and Ls as follows:

$$L = \mu L_p + \lambda L_s$$

Losses are then back propagated to the trainable parameters which are a combination of ResNet-50 encoder, Depth Decoder and Pose Decoder models. The optimizer used is Adam with a StepLR scheduler.

We don't use losses from depth maps to train the network. The depth metrics - rms, squared relative, log_rms, a1, a2, a3 are as proposed [14].

$$a1 = mean(\delta < 1.25)$$
$$a2 = mean(\delta < 1.25^2)$$
$$a3 = mean(\delta < 1.25^3)$$

where,
$$\delta = max(\frac{depth_{gt}}{depth_{pred}}, \frac{depth_{pred}}{depth_{gt}})$$

$$abs\_rel = mean(\frac{|depth_{gt} - depth_{pred}|}{depth_{gt}})$$

$$sq\_rel = mean(\frac{(depth_{gt} - depth_{pred})^2}{depth_{gt}})$$

$$rms = \sqrt{mean((depth_{gt} - depth_{pred})^2)}$$

$$log\_rms = \sqrt{mean((\log_{10} depth_{gt} - \log_{10} depth_{pred})^2)}$$

Next, coming to the hyperparameters, we used a batch size of 12, lr=1e-4, a lr scheduler with steps every 15 batches for a total of 20 epochs. The training completed in approximately 8 hrs. Looking at the results, we were able to replicate the performance metrics as mentioned by the author [6] with an error threshold of 0.5% . A sample prediction is displayed in Fig. 6.

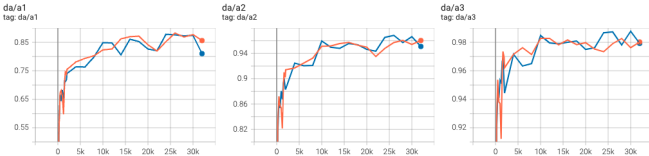

Fig. 4 Depth Accuracy Metrics - a1, a2, a3 calculated for thresholds 1.25, 1.25^2 and 1.25^3 (left to right).

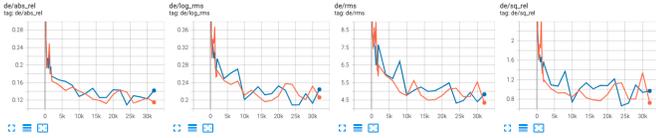

Fig. 5 Depth Error Metrics - abs_rel, log_rms, rms, sql_rel (left to right).

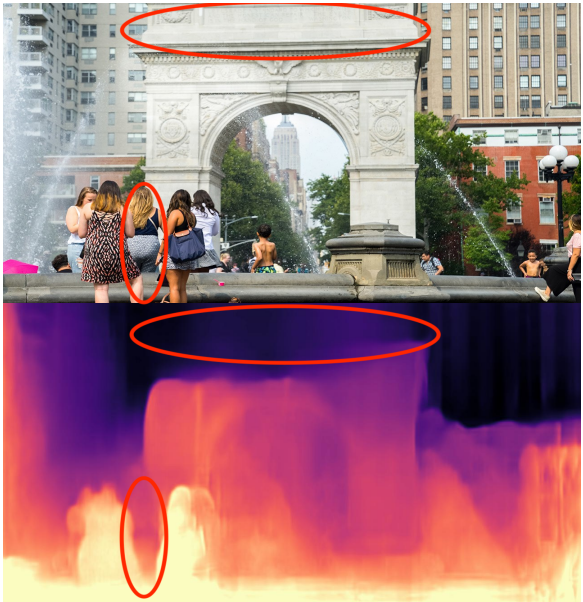

Fig. 6 Sample Prediction - Monodepth2 Impln. Using only encoder and depth decoder to calculate disparity map and further to depth map. Notice the missing depth mask of person and top of WSP.

*D. Building Proposed Architecture - A. Mishra, M. Poddar*

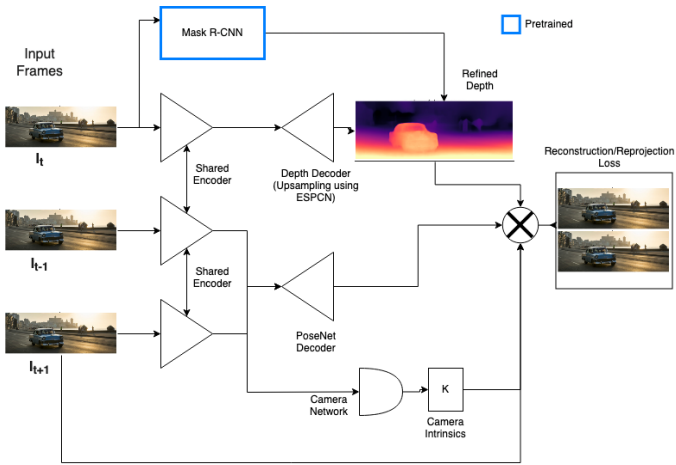

Fig. 7 Proposed Architecture with Monodepth2 Baseline Resnet Encoder, Depth Decoder (with ESPCN Upsampling), Pose Decoder, Mask R-CNN model and CamNet

We have built upon the work of Monodepth2 [6] by adding feature maps from instance-level segmentation masks from Mask R-CNN which is pretrained on the COCO dataset. The idea is to select semantically important objects like person, vehicle, animals and adjust the depth values accordingly. Please note that this is a kind of approximation which might not be completely correct but will provide good results for the majority of content.

We also have integrated a learnable camera intrinsic network from WildNet [21]. With the help of this we will be able to learn camera intrinsics along with depth and pose parameters. The camera network is inspired from [21]. The PoseNet Encoder's feature outputs are delivered to the camera network, followed by a series of processes to decrease the number of channels to 256. Further, this output will be squeezed using adaptive AvgPooling to form a smaller feature input. The normalized focal lengths fx,fy and main offsets cx,cy in both horizontal and vertical axes will then be estimated using two 3x3 convolution layers on the new input. The camera parameters matrix K is created by concatenating these values. The softplus activation function $f(x) = \log(1 + \exp(x))$ is used to avert negative values for focal lengths.

Lastly, we have replaced the bilinear upsampling with ESPCN[9] to improve the propagation of intuitive features during upsampling. This is inspired by the Image Super Resolution technique.

*E. Implementing Proposed Architecture - A. Mishra, M. Poddar, M. Kewlani*

We inherited all the loss metrics and performance benchmark from Monodepth2 for easy comparison between the baseline and our model.

Using a pre-trained Mask R-CNN[23] model, we get the segmentation mask for each object identified in the input image. These masks with confidence >= 0.7 and belonging to certain classes are then merged together to form a semantic map which outlines all the identified objects in the image. This map is then used to mark the identified object clearly in the disparity while retaining the intensity value from the disparity map. This results in a much more coherent map compared to the baseline implementation. Here, we initially tried adjusting the disparity for both losses i.e. smoothness loss and reprojection loss. But later realized that added disparity adjustment for smoothness loss is not really required as it is just calculating the difference in the intensity gradient between original image and disparity map. So we only adjusted the disparity map for reprojection loss which gave us improved performance compared to the initial work.

Similarly, using pytorch Implementation of the paper [21] we got the K and inverse K values which were used to reconstruct our future and past frames. Considering we used nn.Module for creating the intrinsics learning class, we just had to pass the required input into the parameter and further learning and adjustment were automatically done. We leverage bottleneck output which we derive from the existing Pose Decoder model. The bottleneck output is passed to a convolutional block of kernel 1x1 with 2 output channels and softplus activation to get the focal lengths ($f_x$ and $f_y$). Similarly, the bottleneck output is passed to a secondary convolutional block of kernel 1x1 with 2 output channels to get cx and cy offsets.

Lastly, ESPCN [9] upsampling is added to the Depth Decoder Model in the U-Net architecture. The architecture is such that it takes some encoded features of an image with C channels, convolves it a couple of time with kernel size 3 with output channels as C*(r*r), where r is the upscaling factor (in our case, 2). Finally, the channels are redistributed such that the factor r is multiplied in height and width axes and the channels are divided by $r^2$.

Also, We noticed that the KITTI [24] dataset was essentially captured over a couple of weeks where almost all the frames are from sunny weather. Since, self-driving vehicles encounter a lot of variations in the weather, we explored changing the baseline color augmentations to include weather variations. So, we used the albumentations library to facilitate weather augmentations [22]. We used RandomSnow, RandomSunflare, RandomFog and RandomRain weather augmentations with probability of 0.3 each on the dataset.

### III. Results

We have conducted experiments to evaluate the impact of each add on described in the proposed architecture. The results of the experiments are in Table I. We can observe how implementing the proposed architecture has improved the rmse of the model, hence resulting in a better performance than MonoDepth2 [6]. Fig 8 shows the prediction from MonoDepth2 [6] and our model. Our code is available at : https://github.com/nyu-ce-projects/depth-estimation

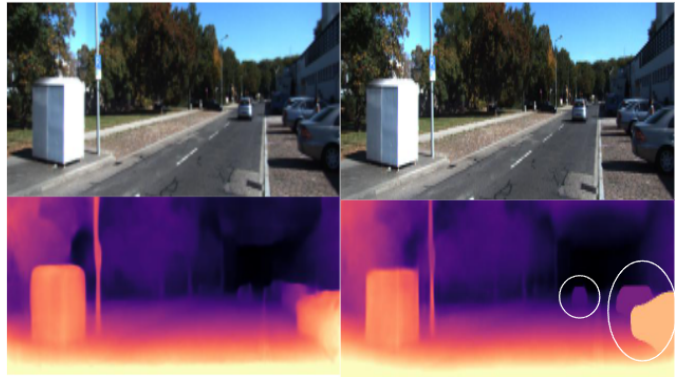

Fig. 8 Monodepth2 and Our Model Prediction Comparison

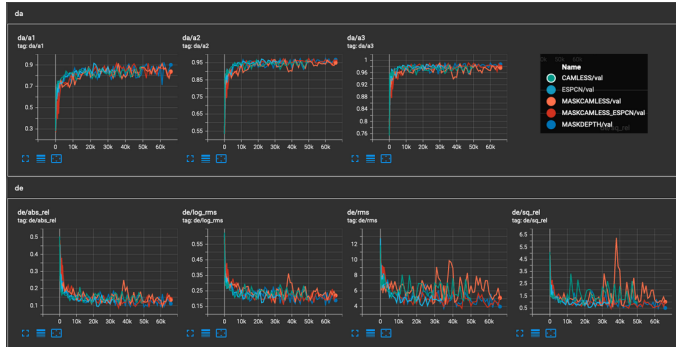

Fig. 9 Training curves on Validation dataset

Since we are considering monocular sequences, the camera rotation will R = 1. Hence reprojection expression will be reduced to z'p' = zp + Kt. Here we can see that loss depends on the expression Kt, so even though K and t might be incorrect, we will get the correct product Kt because there exists a K't' = Kt.

$$\text{Baseline fixed } K = \begin{bmatrix} 0.58 & 0.00 & 0.50 \\ 0.00 & 1.92 & 0.50 \\ 0.00 & 0.00 & 1.00 \end{bmatrix}, \text{ Learned } K = \begin{bmatrix} 0.8730 & 0.0000 & 1.8587 \\ 0.0000 & 0.7521 & 1.2776 \\ 0.0000 & 0.0000 & 1.0000 \end{bmatrix}$$

### IV. Summary

We attempted these 4 additions to the MonoDepth2[6] paper for Self-Supervised Depth Estimation Task: Disparity Enhancement using **Mask R-CNN,** Upsampling using **ESPCN** in Depth Decoder, Learnable Camera Intrinsics (K) using **CameraNet**, Improve generalization using **Weather Augmentations.**

**Mask R-CNN** model leverages benefits of segmentation tasks from COCO dataset to improve coherent inverse depth map refinements. Trade-off - RMS improvement **18.22%**, training time increased by **300%**.

Addition of **ESPCN** for upsampling in UNet Decoder instead of bilinear interpolation at each scale helps the decoder model to not lose encoded features. No significant training time increase was observed with **15.77%** RMS improvement.

Monodepth2 uses t-1, t and t+1 frames as substitutes for left and right images to calculate the pose parameters -

TABLE I
EXPERIMENTATION RESULTS

| Implementation | a1 | a2 | a3 | abs_rel | rms | log_rms | sq_rel |
|---|---|---|---|---|---|---|---|
| MonoDepth2 [6] | 0.877 | 0.959 | 0.981 | 0.115 | 4.863 | 0.193 | 0.903 |
| CamLess[10] | 0.891 | 0.964 | 0.983 | 0.106 | 4.482 | 0.182 | 0.750 |
| **Ours**-MonoDepth2+ Mask R-CNN | 0.9008 | 0.9684 | 0.9872 | 0.1117 | 3.977 | 0.1886 | 0.5114 |
| MonoDepth2+Mask R-CNN + ESPCN | 0.8403 | 0.9651 | 0.9858 | 0.1214 | 4.096 | 0.205 | 0.6251 |
| **Ours** - MonoDepth2 + CamLess | 0.8629 | 0.9542 | 0.98 | 0.1186 | 4.737 | 0.2103 | 0.7843 |
| **Ours** - MonoDepth2 + CamLess+Weather Augmentation | 0.8704 | 0.9582 | 0.9789 | 0.1223 | 4.934 | 0.2016 | 0.9271 |
| **Ours** - MonoDepth2 + Mask R-CNN + CamLess | **0.9148** | 0.9685 | 0.9832 | 0.0996 | 4.25 | 0.1887 | 0.5722 |
| **Ours** - MonoDepth2 + Mask R-CNN + CamLess** | 0.879 | **0.9699** | **0.9876** | 0.111 | 3.959 | **0.177** | 0.5079 |
| **Ours** - MonoDepth2 + Mask R-CNN + ESPCN + CamLess | 0.9105 | 0.9637 | 0.9814 | **0.0956** | 3.746 | 0.1858 | 0.4868 |
| **Ours** - MonoDepth2 + Mask R-CNN + ESPCN + CamLess** | 0.8854 | 0.9621 | 0.9842 | 0.1166 | **3.485** | 0.1884 | **0.4793** |

** Skipping disparity adjustment for smoothening loss

axisangle (transformed to rotation) and translation. Since, this assumption of small motion with fixed K matrix ($f_x$, $f_y$, $c_x$, $c_y$) can hinder training, **CameraNet** generalizes the motion estimation using trainable K. No significant time increase was observed with **28.34%** RMS loss improvement to the original baseline paper.

In order to generalize more real world scenarios, **Weather Augmentations** help the model to generalize better. Although we haven't evaluated our model on videos from the wild yet for a concrete conclusion, we expect it to generalize better on other outdoor datasets like CityScapes.

**Hyperparameters Tuning and Infrastructure Used**:
AdamW Optimizer with lr = 1e-4, weight_decay = 5e-2, batch_size = 12, Scheduler: CosineAnnealingLR with 20 epochs, and 0 minimum lr

## V. FUTURE WORK

We are also working on implementing depth estimation tasks using a vision transformer, and would like to integrate a vision transformer in the self-supervised depth estimation from monocular images, and compare it with the state of the art models. We take guidance from the paper Vision Transformer for Depth Estimation[3], and use the Dense Prediction Transformer instead of Encoders and Decoders. Also,We are looking at the possibility of using EfficientNets[15] or ConvNext[18] instead of the ResNet shared encoder to improve the inference time of our depth estimation models. Our take: Less Trainable Parameters → Less Floating Point Operations → Less Inference Time.